\documentclass[conference]{IEEEtran}
\usepackage{times}
\usepackage{graphicx}

\usepackage[numbers]{natbib}
\usepackage{multicol}
\usepackage[bookmarks=true]{hyperref}
\usepackage{booktabs}

\begin{document}

\title{Using Language and Road Manuals to Inform Map Reconstruction for Autonomous Driving}

\author{
\IEEEauthorblockN{Akshar Tumu\IEEEauthorrefmark{1},
Henrik I. Christensen\IEEEauthorrefmark{1},
Marcell Vazquez-Chanlatte\IEEEauthorrefmark{2},
Chikao Tsuchiya\IEEEauthorrefmark{2}, and
Dhaval Bhanderi\IEEEauthorrefmark{2}}

\IEEEauthorblockA{\IEEEauthorrefmark{1}Department of Computer Science and Engineering, University of California San Diego, La Jolla, CA, USA\\
\texttt{\{atumu, hichristensen\}@ucsd.edu}}
\IEEEauthorblockA{\IEEEauthorrefmark{2}Nissan North America, Santa Clara, CA, USA\\
\texttt{\{Marcell.VazquezChanlatte, Chikao.Tsuchiya, Dhaval.Bhanderi\}@nissan-usa.com}}
}

\maketitle
\renewcommand{\thefootnote}{}
\footnotetext{This research is supported by Nissan.}
\renewcommand{\thefootnote}{\arabic{footnote}}

\begin{abstract}
Lane-topology prediction is a critical component of safe and reliable autonomous navigation. An accurate understanding of the road environment aids this task. We observe that this information often follows conventions encoded in natural language, through design codes that reflect the road structure and road names that capture the road functionality. We augment this information in a lightweight manner to SMERF, a map-prior-based online lane-topology prediction model, by combining structured road metadata from OSM maps and lane-width priors from Road design manuals with the road centerline encodings. We evaluate our method on two geo-diverse complex intersection scenarios. Our method shows improvement in both lane and traffic element detection and their association. We report results using four topology-aware metrics to comprehensively assess the model performance. These results demonstrate the ability of our approach to generalize and scale to diverse topologies and conditions.
\end{abstract}

\IEEEpeerreviewmaketitle

\section{Introduction}
Lane-topology information~\cite{openlanev2} is crucial for Autonomous vehicles to plan and navigate complex environments safely. Lane-topology includes the geometry of the lanes and the associations of the lanes between themselves and with other traffic elements (e.g., traffic lights and signs).

High Definition (HD) Maps are a reliable source for detailed lane-topology information. However, their high maintenance cost and limited geographical coverage make them difficult to scale for real-world deployment. Recent works such as SMERF \cite{smerf} show that Standard Definition (SD) Maps, such as OpenStreetMap (OSM) \cite{osm}, provide valuable road-level priors which, when combined with onboard sensor data, can be used to predict the detailed lane-topology of a road.

A key limitation of these works is that they generally only use road geometry and the road-type labels (e.g, highway=residential) from OSM while ignoring the other semantic and geometric information available for each road. Lane-level cues, such as the number of lanes and lane width, are also overlooked, even though they could aid lane-topology prediction in scenarios with camera occlusion, unmarked roads, or complex road intersections. 

Notably, the missing information mirrors how the built world is constructed and named through the design guidelines implicitly or explicitly encoded in language. For example, even if the road type is unlabeled in OSM, names such as ``Main Street'', ``Patricia Circle'', or ``Oregon Expressway'' act as strong cues. Similarly, many locales require infrastructure to be built according to design codes written in natural language, diagrams, and tables. 
We hypothesize that capturing these cues can accelerate the deployment and adaptation of topology prediction systems to diverse geographies.

\emph{Our key insight is to use
language embeddings and Retrieval Augmented Generation} to provide these cues to an existing SD-Map informed HD-map reconstruction model. We evaluate our method on two geographically distinct locations and show improvement across four metrics capturing different aspects of lane-topology, highlighting our method's generalization and real-world applicability.

\section{Related Work}
\subsection{High Definition Maps}
HD Maps consist of detailed lane-level information and annotations. They are used in various downstream Autonomous Driving tasks such as Trajectory Prediction (\cite{hivt}, \cite{densetnt}), Path Planning (\cite{hdm-rrt}, \cite{planning2}), etc., since they have detailed lane-level and traffic element annotations. Popular sources of HD Maps include nuScenes~\cite{nuscenes}, Argoverse 2~\cite{argoverse2}, Waymo Open Dataset~\cite{waymo}, etc. However, it took several months to build and annotate these HD Maps. Another issue is that these HD Maps are restricted to a few geographical areas, limiting their global scalability. For example, nuScenes only has maps for Boston, USA, and Singapore, while Argoverse 2 only covers six cities in the United States.

\subsection{Online Lane-Topology Prediction}
Online lane-topology prediction models (\cite{hdmapnet}, \cite{vectormapnet}, \cite{maptrv2}) utilize the data collected by onboard sensors, such as cameras, to predict the lane-topology of the roads on the fly. 

Recent works such as SMERF~\cite{smerf}, TopoSD~\cite{toposd}, and ImagineMap~\cite{imaginemap} introduce SD Map as a road-level prior to help online lane-topology models do long-horizon prediction or predict lane-topology in the presence of camera occlusion. While SD Maps don't contain lane-level details, they consist of the road geometry. SMERF, in particular, encodes road centerline geometry (as polylines) and road type labels from OSM using a Transformer module and then performs cross-attention between the BEV-representation of the multi-camera features and the map features. This is followed by a decoder head (Deformable-DeTR~\cite{deformabledetr}) which outputs the lane centerlines and the topology.

SD Maps like OSM provide the advantage of being available open-source and for a major portion of the world. OSM also provides road metadata such as the number of lanes, service classification of the road, road direction, and speed limit, etc., which has crucial semantic information captured in natural language. This information can help disambiguate road functionality and infer lane connectivity. One key detail that is very inconsistently tagged in OSM is the lane width of a road. Lane widths can help deduce the spatial layout of multi-lane roads even in the absence of clear lane markings. \citet{sd++} show that the standardized road design manuals, such as the Highway Design Manual~\cite{hdm}, provide the lane widths for different road types, which can be used for lane-topology prediction. 

Another class of methods (\cite{nmp}, \cite{streammapnet}, \cite{uniprevpredmap}) learn temporal priors from past traversals to incrementally improve the map. However, they require more training, and the priors don't generalize geographically. In contrast, we inject structural priors from globally available SD Maps and road design manuals, making it scalable.

\subsection{Case Study: Real-world applicability of Online Mapping Methods}\label{subsec:case study}
We discuss the real-world applicability of Online Mapping Methods by taking the example of the ‘Trajectory Prediction' downstream task. Trajectory Prediction involves predicting the trajectory of a vehicle for a few seconds into the future. Trajectory prediction models rely on HD Maps for detailed lane geometry and semantics since it improves the accuracy and reliability of the predicted vehicle paths.

Two particular works - \citet{traj1}, \citet{traj2} evaluate the performance of DenseTNT \cite{densetnt}, a popular Trajectory Prediction model, when trained on HD Maps and maps generated by Online Lane-Topology Prediction models. The lane-topology prediction models evaluated are variants of MapTRv2~\cite{maptrv2} and StreamMapNet~\cite{streammapnet}. The models train on two seconds of annotated vehicle trajectories and ego-vehicle pose from nuScenes, along with either online-predicted or ground-truth nuScenes HD maps, to predict vehicle trajectories three seconds into the future. The evaluation metrics (minADE, minFDE, Miss Rate) analyze the alignment of the predicted trajectories with the real-world motion of the vehicles.

These works report that while the StreamMapNet-based model performs very closely to the HD Map-based model, the model that uses the lane-centerline-predicting variant of MapTRv2 even outperforms the HD Map-based model. Such results demonstrate the viability of online lane-topology prediction methods as alternatives to HD Maps for downstream Autonomous Vehicles tasks.

\section{Methodology}
\subsection{Obtaining OSM Metadata}
Each road in an OSM Map has multiple annotated metadata fields such as ‘name', ‘geometry', ‘highway', ‘lanes', etc. The map embedding generated by SMERF's map encoder consists of the ‘geometry' (road polylines) and ‘highway' fields. Now, we extract all other fields annotated for each road apart from ‘osmid', ‘name', and the fields already present in the graph embeddings. While we exclude the ‘name' field, we extract the road suffix (Avenue, Street, etc.) from it, if present. Some of the fields we capture are - ‘service', ‘oneway', ‘number of lanes', ‘maxspeed', etc. We then concatenate the fields into a string and embed them using OpenAI's text-embedding-3-small model\footnote{\url{https://platform.openai.com/docs/models/text-embedding-3-small}}.

\subsection{Extracting Lane Widths}
To extract lane widths from the Road design manual, we follow the Retrieval Augmented Generation pipeline setup by \citet{sd++} with the Llama3.3 LLM \cite{llama}. We first start by splitting the Highway Design Manual into chunks and embedding them using an embedding model. These embeddings are stored in a vector database.

\begin{figure}[!ht]
    \centering
    \includegraphics[width=\linewidth]{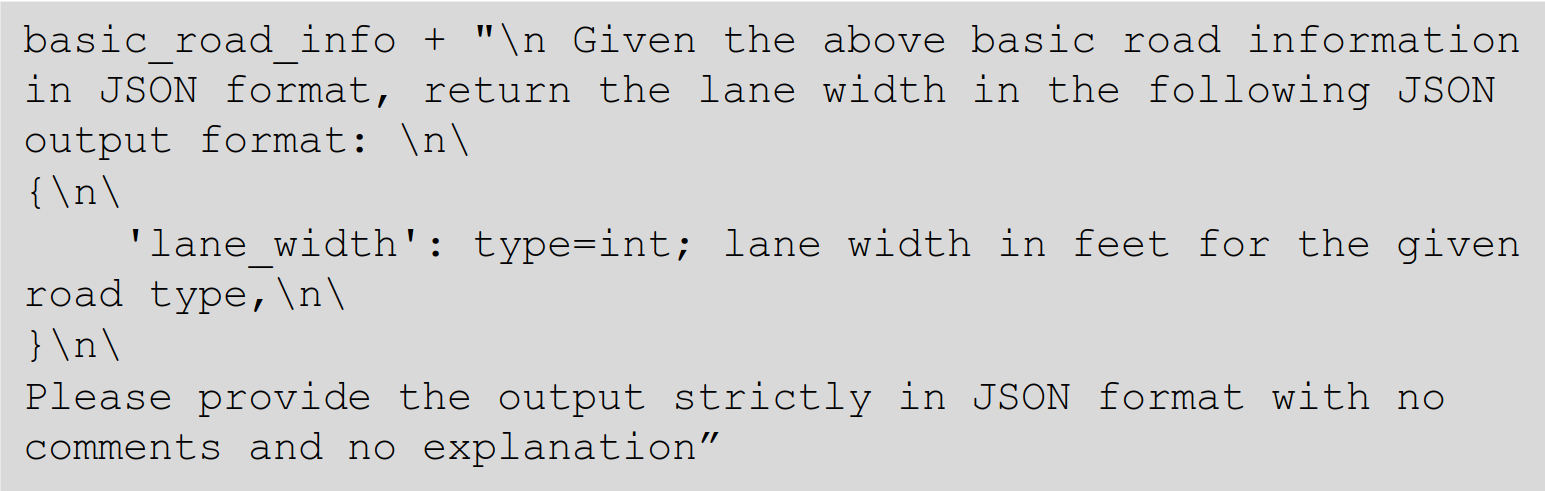}
    \caption{RAG Prompt}
    \label{fig:prompt}
\end{figure}
Figure \ref{fig:prompt} shows the prompt that we query the LLM with. For each road, we also pass its ‘basic\_road\_info' consisting of the fields extracted as part of the OSM metadata. When queried for each road in a map individually, the LLM outputs the corresponding lane widths by retrieving and reasoning upon relevant parts of the road design manual. We embed the lane widths using OpenAI's text-embedding-3-small model.

\subsection{Passing Embeddings to SMERF}
The embedded OSM metadata and Lane Widths are added since they are of the same dimension. The combined text embedding is passed to SMERF's BEV Constructor module, which also takes the Map embeddings as input. The text embeddings are first transformed into the embedding space of the map embeddings using a 2-layer MLP. Then, the Text and Map embeddings are added. Mathematically, the addition is defined as follows:
\begin{equation}\label{eq: 1}
e_p = G(p) + MLP(t_p)
\end{equation}
Where $p$ stands for the polyline of a road, $G(p)$ stands for the graph embedding of that polyline, and $t_p$ stands for the text embedding of that road.

The rest of the pipeline is the same as SMERF, which involves cross-attention between the camera BEV features and the combined embeddings, followed by a decoder to generate the vectorized HD Map output.

\section{Experimental Setting and Evaluation}
\subsection{Dataset}
\begin{figure*}[!htb]
    \centering
    \begin{minipage}[t]{0.3\textwidth}
        \centering
        \includegraphics[height=\linewidth]{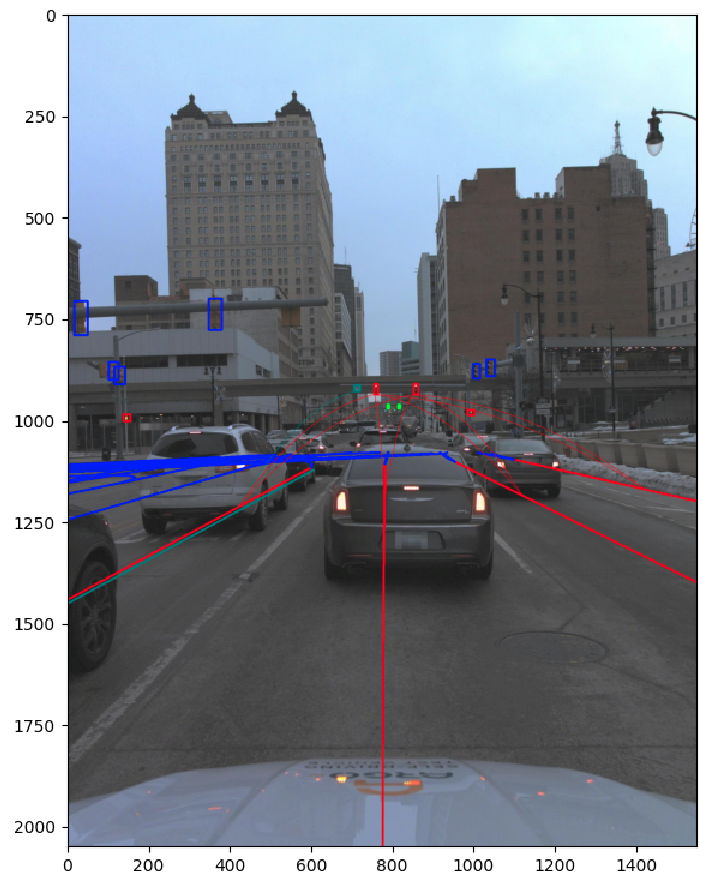} \\
        {\small (a) Perspective view - Detroit}
    \end{minipage}
    \hfill
    \begin{minipage}[t]{0.3\textwidth}
        \centering
        \includegraphics[height=\linewidth]{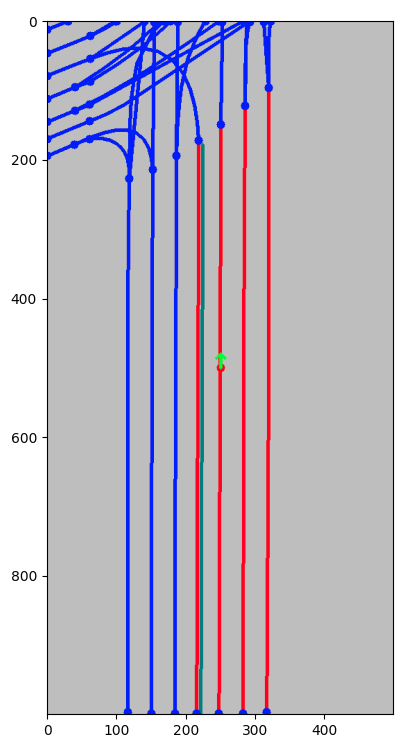} \\
        {\small (b) BEV Lane-topology - Detroit}
    \end{minipage}
    \hfill
    \begin{minipage}[t]{0.3\textwidth}
        \centering
        \includegraphics[height=\linewidth]{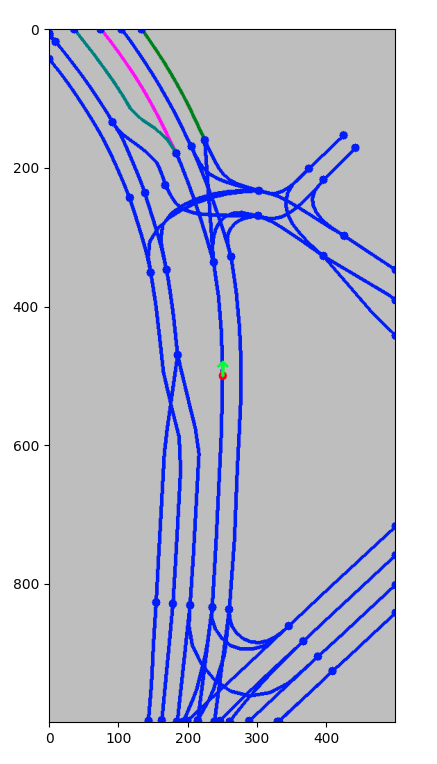} \\
        {\small (c) BEV Lane-topology - Pittsburgh}
    \end{minipage}
    \caption{Comparison of the front-view image and corresponding Ground-Truth BEV lane-topologies. Lane colors in the BEV Map indicate intended actions at intersections, such as left turn (blue), right turn (green), etc. The red dot and the green arrow jointly indicate the ego-pose of the vehicle.}
    \label{fig:three_views}
\end{figure*}
We select two complex road scenarios from the OpenLane-V2 dataset, which are from different cities in the United States, for training and evaluating our model. The two scenarios have 64 frames in total. Figure \ref{fig:three_views} illustrates the Front-view and BEV Lane-topologies of the scenarios with annotated lanes and traffic elements. The first scenario is from Detroit, which involves an upcoming intersection that is occluded in the camera. The second scenario is from Pittsburgh, which involves a complex multi-lane merging scenario.

\subsection{Evaluation Metrics}
We select the following evaluation metrics from the OpenLane-V2 dataset:
\begin{itemize}
\item $DET_l$: Average precision of 3D lane centerline detection using Fréchet distance.
\item $DET_t$: Average precision of traffic element detection across different detected attributes of the traffic elements.
\item $TOP_{ll}$: Accuracy of the predicted connectivity among the detected lane centerlines.
\item $TOP_{lt}$: Accuracy of the predicted association between lane centerlines and traffic elements.
\end{itemize}
We also include a fifth metric, OpenLane-V2 Score (OLS), which is the weighted average of all four metrics.

These metrics evaluate the spatial accuracy of the models and their ability to comprehend traffic elements and associate them with the lanes, thus covering all lane-topology aspects for any complex road scenario in the world.
\begin{table*}[!ht]
    \centering
    \caption{Performance comparison of different model and fusion configurations. ‘$\lambda$' configuration stands for the embedding fusion defined in Equation \ref{eq: 2}. ‘RAG' configuration stands for the combined text embedding defined in Equation \ref{eq: 3}.}
    \renewcommand{\arraystretch}{1.2}
    \resizebox{0.75\textwidth}{!}{%
    \begin{tabular}{l c c c c c c c}
        \toprule
        Model & $\lambda$ & \#Trainable Params & DET$_l$ & DET$_t$ & TOP$_{ll}$ & TOP$_{lt}$ & OLS \\
        \midrule
        SMERF \cite{smerf} & - & 39.62M & 0.5364 & \underline{0.9734} & 0.0595 & 0.3645 & 0.5894 \\
        \hline
        F0 & - & 132K & 0.4134 & 0.9392 & 0.0141 & 0.2125 & 0.4831 \\
        F0 + $\lambda$ & 0.869 & 132K & 0.4138 & 0.9392 & 0.0137 & 0.2122 & 0.4827 \\
        F0 + RAG & - & 132K & 0.414 & 0.9392 & 0.0141 & 0.2123 & 0.4831 \\
        \hline
        F1 & - & 26.28M & 0.5267 & 0.9703 & 0.0538 & 0.3519 & 0.5806 \\
        F1 + $\lambda$ & 0.869 & 26.28M & \underline{0.5391} & 0.9726 & 0.0583 & 0.3462 & 0.5854 \\
        F1 + RAG & - & 26.28M & \textbf{0.5468} & 0.9725 & 0.0628 & 0.3692 & \underline{0.5944} \\
        \hline
        NF & - & 39.75M & 0.5327 & 0.9652 & \textbf{0.0691} & 0.3452 & 0.5871 \\
        NF + $\lambda$ & 0.869 & 39.75M & \underline{0.5391} & \textbf{0.9745} & \underline{0.0674} & \textbf{0.3905} & \textbf{0.5995} \\
        NF + RAG & - & 39.75M & 0.5376 & 0.9671 & 0.0669 & \underline{0.3748} & 0.5939 \\
        \bottomrule
    \end{tabular}
    }
    \label{tab:performance_comparison}
\end{table*}

\subsection{Training and Fusion Configurations}
\subsubsection{Model Training Configurations}
We first start by training the original SMERF model for 100 epochs. Then we further train the model for a second round of 100 epochs by including the OSM Metadata embeddings and Lane-width embeddings. We test the following three training configurations:

\begin{itemize}
\item F0: We freeze the entire model, except the 2-layer MLP for transforming the text embeddings, for the second round of training.
\item F1: We only freeze the multi-camera BEV encoder and the map encoder for the second round of training.
\item NF: No part of the model is frozen for the second round of training.
\end{itemize}

\subsubsection{Embedding Fusion Configurations}
Equation \ref{eq: 1} defines the fusion of the graph and text embeddings for a lane. Our first fusion strategy consists of only the OSM Metadata embedding as the text embedding ($t_p = o_p$ where $o_p$ stands for the OSM metadata embedding).

Our second fusion strategy consists of a weighted addition scheme defined below:
\begin{equation}\label{eq: 2}
e_p = G(p) + \lambda \cdot MLP(t_p) \quad \mbox{where} \quad t_p = o_p
\end{equation}
We set $\lambda$ to be a learnable parameter by the network.

Our third fusion strategy consists of fusing the OSM Metadata embeddings with the Lane-width embeddings obtained from the RAG-LLM system. Mathematically, we define the fusion as follows:
\begin{equation}\label{eq: 3}
t_p = o_p + l_p
\end{equation}
where $l_p$ stands for the Lane-width embedding.

\section{Results}
Table \ref{tab:performance_comparison} shows the results over the two selected scenarios for the baseline (SMERF) and our different model configurations. For the baseline, we use the original SMERF model trained for 100 epochs as our loading checkpoint and train the model for 100 more epochs.

We can observe that our best configurations outperform the baseline for all metrics. Among F0, F1, and NF training configurations, NF has the best results for four of the five metrics, with ‘NF + $\lambda$' variant performing the best for three of the five metrics. However, our ‘F1 + RAG' variant gives the best lane centerline detection performance. F0 variants underperform since all other variants have a lot more trainable parameters. 

Among all four metrics, we see the maximum improvement in the $TOP_{ll}$ metric by 16\%, followed by $TOP_{lt}$ by 7\%. This is because the OSM Metadata embeddings consist of information about the functional type and semantic properties of the lanes, which aid lane connectivity and association between lanes and traffic elements.

\begin{figure}[!htb]
    \centering
    \begin{minipage}[t]{0.24\textwidth}
        \centering
        \includegraphics[height=\linewidth]{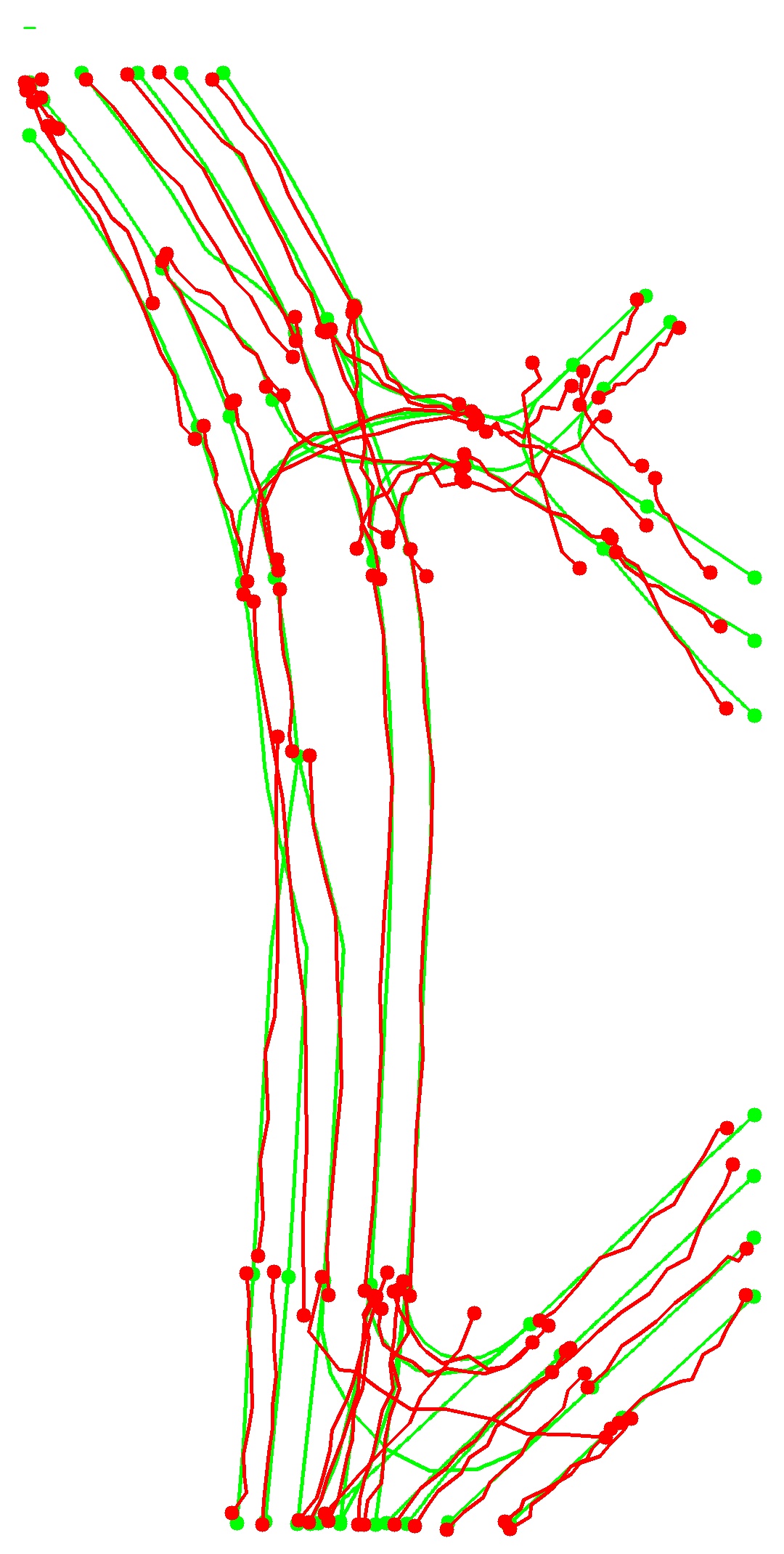} \\
        {\small (a) Pittsburgh}
    \end{minipage}
    \begin{minipage}[t]{0.24\textwidth}
        \centering
        \includegraphics[height=\linewidth]{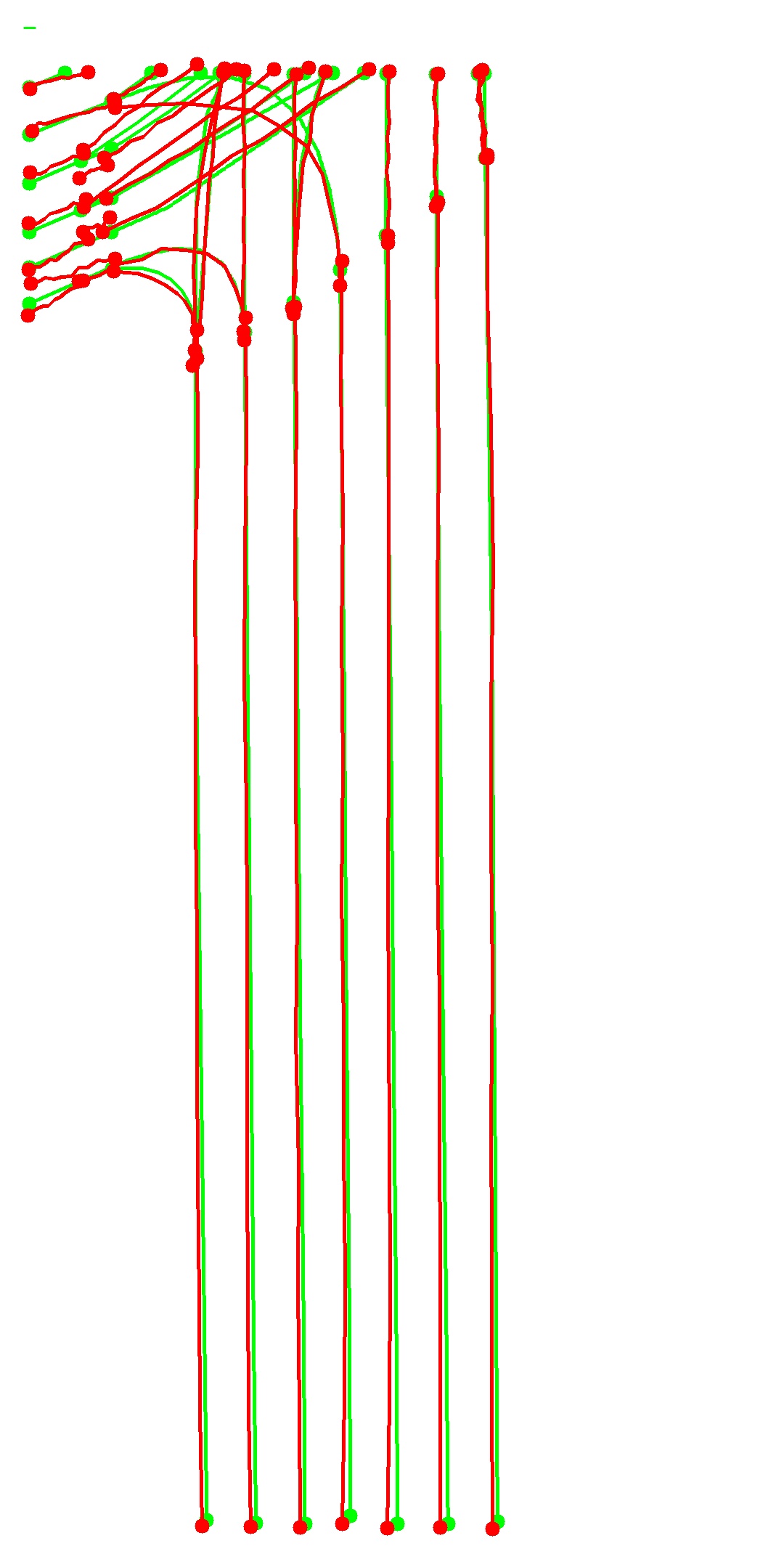} \\
        {\small (b) Detroit}
    \end{minipage}
    \caption{Lane-topology predictions in the BEV view for the two scenarios. The green lines correspond to the Ground Truth lanes, and the red lines correspond to the Predicted lanes. The dots correspond to the joining points between lanes.}
    \label{fig:output}
\end{figure}
Figure \ref{fig:output} shows the lane-topology predictions in BEV view by the ‘NF + $\lambda$' variant of our approach. We can observe that for the Pittsburgh map, there is a slight deviation between the ground truth and predicted lanes, especially at the joining points of the lanes. In contrast, the predictions closely match the ground truth lanes in Detroit's structured map. This highlights the general challenges in mapping curves and intersections compared to structured lanes.
 
\section{Real-World Deployability and Generalization Discussion}
The real-world deployability of the online lane-topology prediction methods has been discussed in Section \ref{subsec:case study}, where the models relying on these techniques match or even outperform the models that use manually annotated HD Maps for a downstream Autonomous Vehicles task.

The qualitative and quantitative results for our approach suggest that integrating multi-camera features with an SD-Map prior and the linguistic priors provided by the SD Maps for each road helps generalize online lane-topology prediction (or, online HD Map building) to even those geometric locations in the world for which manual HD Maps haven't been designed. The evaluation setup, consisting of scenarios from different geometric regions, confirms the generalization ability of our method. 

\section{Conclusion}
We introduce a lightweight, deployable augmentation to SMERF, which integrates semantic and geometric metadata presented in natural language in SD Maps and lane-width priors extracted from road design manuals using RAG to improve online lane-topology prediction. By encoding the said information with an LLM and combining it with the road polyline encodings from SD Maps, our model enhances the SD Maps without modifying the model architecture or using HD Maps.

Our evaluation spans two geographically diverse real-world intersection scenarios from the OpenLane-V2 dataset. Through carefully-chosen topology-aware metrics, we show that our method achieves consistent improvement in all aspects of lane-topology prediction for the chosen complex scenarios. These results demonstrate the practical viability of integrating lightweight, globally available priors into robot perception systems. By operating without the expensive and geometrically constrained HD Maps and handling scenarios with camera occlusion, our method shows scalability and generalizability across geographical regions and conditions.

\bibliographystyle{plainnat}
\bibliography{references}

\end{document}